%% file: main.tex
\documentclass{bmvc2k}

\usepackage{multirow}
\usepackage{graphicx} 
\usepackage{amsmath}
\usepackage{amssymb}

\title{Q-Align: Alleviating Attention Leakage\\in Zero-Shot Appearance Transfer\\via Query-Query Alignment}

\addauthor{Namu Kim\textsuperscript{*}}{namu.kim@kt.com}{1}
\addauthor{Wobin Kweon\textsuperscript{*}}{wonbin@illinois.edu}{2}
\addauthor{Minsoo Kim\textsuperscript{*}}{km19809@postech.ac.kr}{3}
\addauthor{Hwanjo Yu\textsuperscript{\dag}}{hwanjoyu@postech.ac.kr}{3}

\newcommand{\proposed}{Q-Align\xspace}
\newcommand{\Task}{Appearance transfer\xspace}
\newcommand{\task}{appearance transfer\xspace}

\addinstitution{
 KT Corporation, South Korea
}
\addinstitution{
University of Illinois Urbana-Champaign, USA
}
\addinstitution{
Pohang University of Science and Technology (POSTECH), South Korea
}

\runninghead{N. Kim \etal}{Q-Align}

\def\eg{\emph{e.g}\bmvaOneDot}

\def\etal{\emph{et al}\bmvaOneDot}

\begin{document}

\maketitle
\begingroup
\renewcommand\thefootnote{}
\footnotetext{\textsuperscript{*}\;These authors contributed equally to this work.}
\footnotetext{\textsuperscript{\dag}\;Corresponding author.}
\endgroup

\input{sections/0_abstract}
\input{sections/1_intro}
\input{sections/2_attention_leakage}
\input{sections/3_q_align}

\input{sections/4_experiment}
\input{sections/5_related}

\input{sections/6_conclusion}
\input{sections/7_acknowledgement}

\bibliography{egbib}
\end{document}

%% file: sections/0_abstract.tex
\begin{abstract}
We observe that zero-shot \task with large-scale image generation models faces a significant challenge: \textit{Attention Leakage}.
This challenge arises when the semantic mapping between two images is captured by the Query-Key alignment.
To tackle this issue, we introduce \textbf{\proposed}, utilizing Query-Query alignment to mitigate attention leakage and improve the semantic alignment in zero-shot \task.
\proposed incorporates three core contributions: 
(1) Query-Query alignment, facilitating the sophisticated spatial semantic mapping between two images; 
(2) Key-Value rearrangement, enhancing feature correspondence through realignment; and 
(3) Attention refinement using rearranged keys and values to maintain semantic consistency.
We validate the effectiveness of \proposed through extensive experiments and analysis, and \proposed outperforms state-of-the-art methods in appearance fidelity while maintaining competitive structure preservation.
\end{abstract}

%% file: sections/1_intro.tex
\begin{figure}[h!]
    \centering  
    \includegraphics[width=0.85\linewidth]{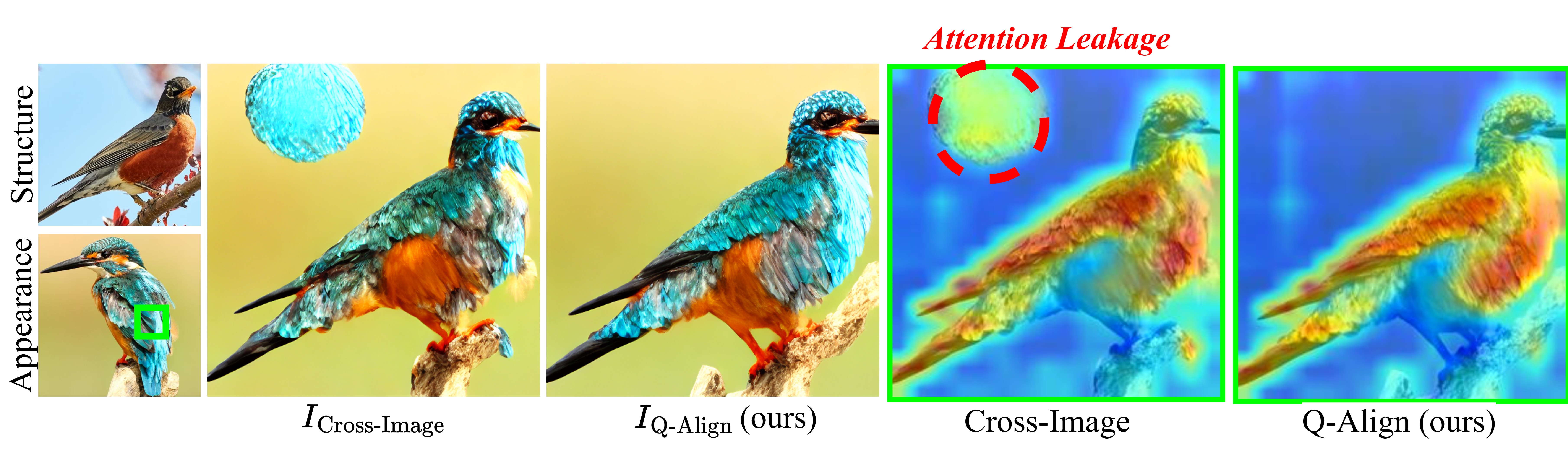}    
    \caption{Attention leakage in Cross-Image~\cite{cross-image}. The green-outlined images show the attention map corresponding to the bird's body (a green square in Appearance image).} 
    \label{fig:intro_leakage_figure}
    \vspace{-0.5cm}
\end{figure}

\section{Introduction}
\Task is an image generation process where the structure image establishes the framework of semantic regions, and the appearance image provides the texture and identity details to fill those regions.
As illustrated in Figure~\ref{fig:intro_leakage_figure}, for example, the wings in the appearance image are transferred to the wing regions of the structure image, branches to branches, and background to background.
To achieve this, the primary challenge is achieving precise semantic alignment between the features of the two images.

Early methods relied on rule-based techniques~\cite{SIFT,SURF} for semantic correspondence, but were limited in their adaptability to diverse and complex datasets.
With advances in deep learning, data-driven approaches~\cite{patchmatch, cat, gocor, raft} were introduced to improve alignment accuracy.
Although these methods demonstrated promising performance, they depend on ground-truth correspondence and require extensive labeled training datasets.
Pre-trained vision transformer features~\cite{dino, asic} have gained attention as a solution to overcome these limitations by enabling the extraction of correspondence maps without labeled datasets~\cite{unsupervised_semantic, splice}.

More recently, large-scale image generation models~\cite{stablediffusion,vqdiffusion,dalle} have introduced powerful generative capabilities that can be leveraged to discover semantic maps~\cite{DIFT, diffusion_semantic_hyperfeature, diffusion_semanticunsupervised, diffusion_semantic_tale_two_feature} in a \textit{zero-shot} manner.
These models are utilized in various ways, such as using the generative model as an encoder to extract semantically aligned features~\cite{DIFT, diffusion_semantic_hyperfeature,diffusion_semantic_tale_two_feature} or through attention control~\cite{masactrl,p2p,pix2pix_zero,plug_and_play} to manipulate specific regions.
Further research has analyzed the role of attention~\cite{attention_analysis} and optimized its mechanisms~\cite{DPL} for improved performance.

The prominent method in zero-shot \task, Cross-Image~\cite{cross-image}, effectively performs appearance transfer without training by leveraging large-scale image generation models and employing attention control, based on MasaCtrl~\cite{masactrl}.
Despite its effectiveness, Cross-Image exhibits a significant challenge: \textit{attention leakage}.
We define the attention leakage as instances where the object(or background) is mapped to the background(or object), or the semantic mapping within the object itself is incorrect. 
As shown in Figure~\ref{fig:intro_leakage_figure}, attention to the bird's body spills over to random background regions rather than remaining focused on the bird itself.

This paper proposes \textbf{\proposed}, a novel approach for zero-shot \task that mitigates attention leakage without requiring any optimization process or additional data.
We provide an in-depth analysis of attention leakage in Cross-Image and introduce a novel approach based on query-query alignment, as follows:
\begin{itemize}
    \item We highlight our key observation that naive mixing of query, key, and value can cause misalignment between features, leading to attention leakage (Section~\ref{sec:Motivation}).
    \item We demonstrate that query-query alignment offers more sophisticated spatial semantic mapping, and propose a key-value
    rearrangement technique to improve query-key-value correspondence in the conventional Cross-Image attention approach (Section~\ref{sec:Q-Align}).
    \item We validate the effectiveness of the proposed method through a novel evaluation protocol leveraging GPT-4o series, and provide a detailed analysis of each component (Section~\ref{sec:experiment}).
\end{itemize}
The official implementation of Q-Align is publicly available\footnote{\url{https://github.com/SEED-TO-TREE/Q-Align-BMVC2025}}.

%% file: sections/2_attention_leakage.tex
\section{Preliminary}
\label{sec:Preliminary}

\textbf{Problem Formulation.}
Given a pair of images, \( I_{\text{app}} \) and \( I_{\text{str}} \), the \task task aims to generate \( I_{\text{out}} \) by transferring the appearance of \( I_{\text{app}} \) onto \( I_{\text{str}} \), mapping semantically corresponding regions and ensuring coherence between objects and backgrounds.
In Figure~\ref{fig:intro_leakage_figure}, features such as wings, beak, and branches in \( I_{\text{app}} \) should be aligned with semantically corresponding regions in \( I_{\text{str}} \). 
Additionally, the background of \( I_{\text{app}} \) needs to be smoothly integrated into that of \( I_{\text{str}} \), creating a cohesive and visually seamless result.

\vspace{0.1cm}
\noindent \textbf{Cross-Image Attention.}
The cross-image attention mechanism \cite{cross-image} is built upon the mutual attention of MasaCtrl~\cite{masactrl}, which demonstrates how the appearance of input image can be semantically mapped to the structure of the output image.
Since self-attention layers allow the model to capture feature correspondences within an image, the queries, keys, and values of these layers can also be leveraged to infer corresponding regions between different images.
Let \((Q_{\text{app}}, K_{\text{app}}, V_{\text{app}})\) and \((Q_{\text{out}}, K_{\text{out}}, V_{\text{out}})\) denote the queries, keys, and values corresponding to \( I_{\text{app}} \) and \( I_{\text{out}} \), respectively, within a specific self-attention layer.
The cross-image attention is defined by replacing $K_{\text{out}}, V_{\text{out}}$ with $K_{\text{app}}, V_{\text{app}}$ as follows:
\begin{equation}
    \text{softmax} \left( \frac{Q_{\text{out}} K_{\text{app}}^\top}{\sqrt{d}} \right) V_{\text{app}}.
\label{eq:attention_app}
\end{equation}
$Q_{\text{out}}$ determines the spatial semantics and is initialized as $Q_{\text{str}}$ at the first diffusion step.
$K_{\text{app}}$ offers appearance context for each query. 
By doing so, Cross-Image enables the model to implicitly transfer the visual appearance between semantically similar objects in the different images.

\section{Motivation}
\label{sec:Motivation}


\textbf{Attention Leakage.}  
Figure~\ref{fig:intro_leakage_figure} shows attention leakage in Cross-Image during the \task from \( I_{\text{app}} \) to \( I_{\text{str}} \), causing unreliable generation in \( I_{\text{Cross-Image}} \).  
To analyze the artifact in the upper left of \( I_{\text{Cross-Image}} \), we examine the attention map for the bird’s body (green square) in \( I_{\text{app}} \).  
The attention erroneously shifts to a background region instead of focusing on the bird (\textit{i.e.}, it \textit{leaks} into irrelevant areas).  
This misaligns the background and draws attention to unrelated features (\eg, a mint-colored bird).  
In our experiment, attention leakage occurs in 94 of 180 pairs (\textbf{52.2\%}), and Q-Align improves 65 of these (\textbf{69.1\%}).

\begin{figure}[t]
\centering
\begin{minipage}[t]{0.4\textwidth}
    \centering
    \includegraphics[width=\textwidth]{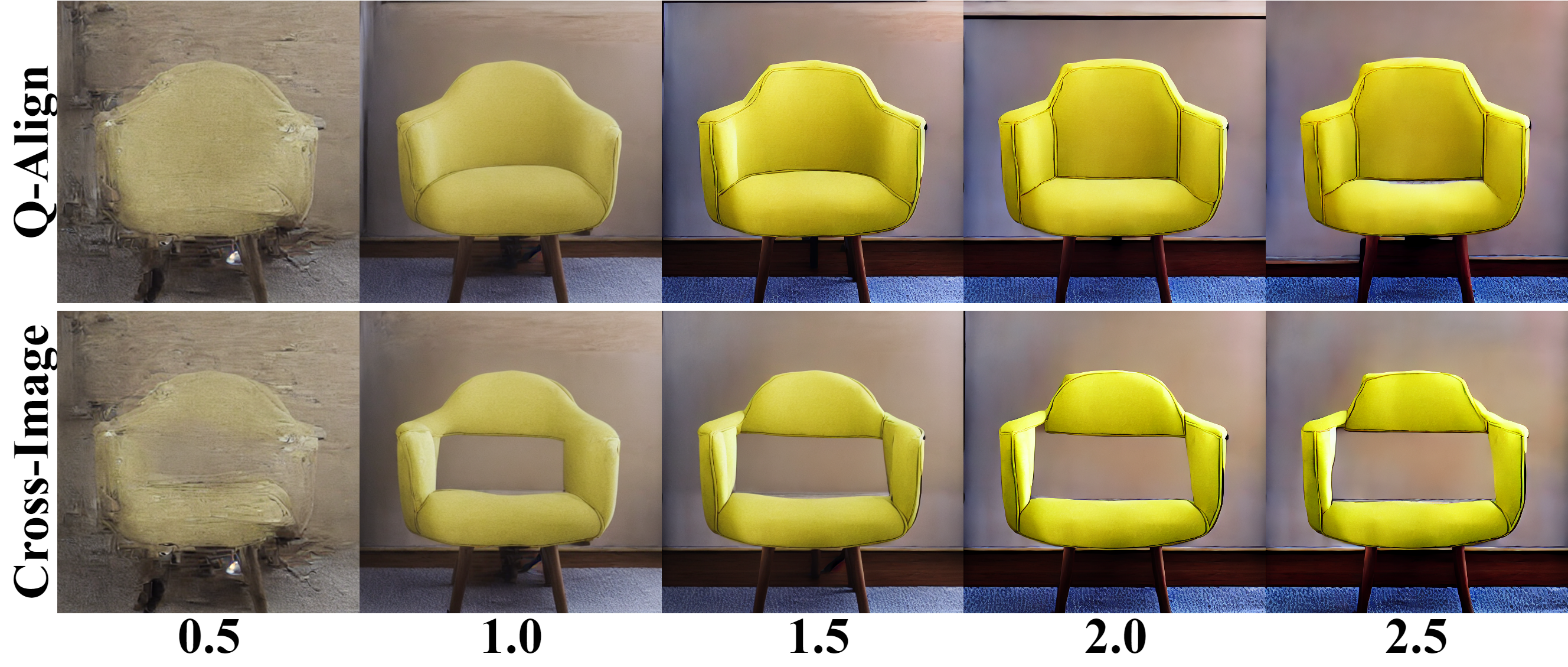}
    \caption{Results of Q-Align (ours) and Cross-Image with varying contrast strength.}
    \label{fig:hyperparameter}
\end{minipage}
\hfill
\begin{minipage}[t]{0.55\textwidth}
    \centering
    \includegraphics[width=\textwidth]{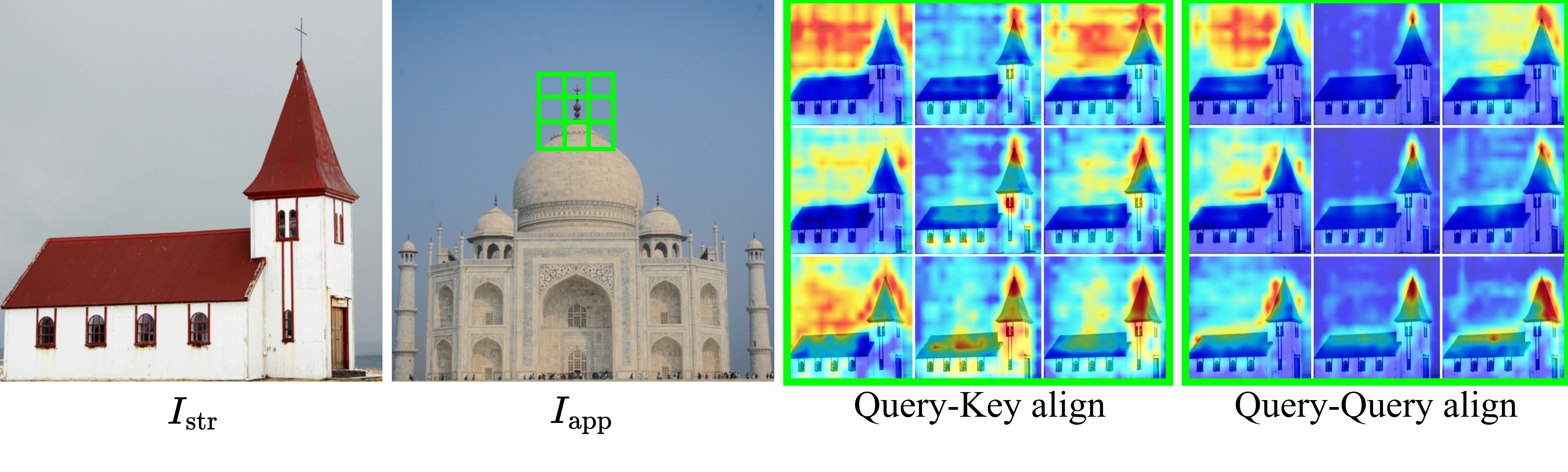}
    \caption{Query-Key vs Query-Query alignment. The right two images present zoomed-in attention maps for the green 3x3 grid in \( I_{\text{app}} \).}
    \label{fig:attention_maps_comparison}
\end{minipage}
\end{figure}

\vspace{0.1cm}
\noindent  \textbf{Query-Key Misalignment.}
Cross-Image relies solely on \( Q_{\text{out}} K_{\text{app}}^\top \) to infer semantically corresponding regions between \( I_{\text{out}} \) and \( I_{\text{app}} \), and faces the attention leakage.
This issue arises when the attention mechanism fails to accurately attend to the corresponding regions, shifting instead to incorrect locations and generating erroneous images.
We will discuss further on Query-Key misalignment with Figure~\ref{fig:attention_maps_comparison} in the following section.



\vspace{0.1cm}
\noindent \textbf{Naive Approaches.}  
Cross-Image introduces an \textit{attention contrasting} hyperparameter to promote relevant alignment and suppress noise.  
One might expect tuning it could fix leakage, but as shown in Figure~\ref{fig:hyperparameter}, the issue persists without proper Query-Key alignment.  
Another strategy is to apply an object \textit{mask} \cite{dragondiffusion}, yet leakage can still occur within the masked area.  
Even with accurate masks, the model often overfits local style rather than ensuring semantic alignment, resulting in unnatural images.  
We revisit this in the experiments (Figure~\ref{fig:qualitative}).



%% file: sections/3_q_align.tex
\section{\proposed}\label{sec:Q-Align}

\begin{figure*}[t]
    \centering  
    \includegraphics[width=0.9\linewidth]{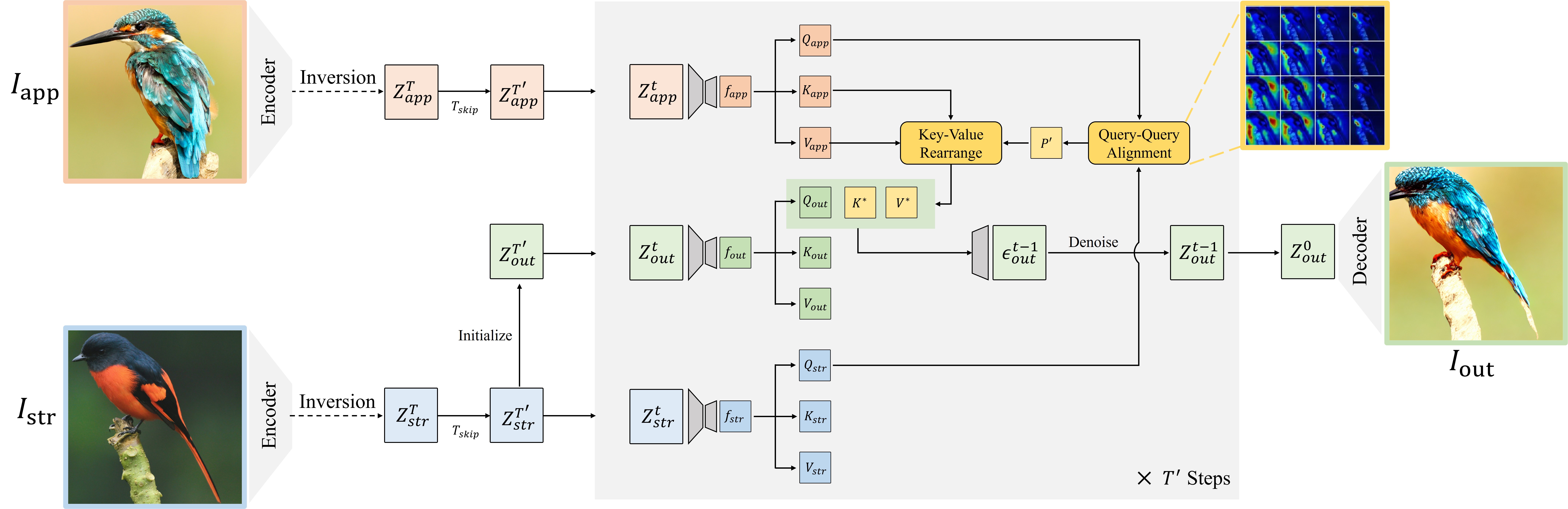}    
    \caption{Illustration of \proposed. We omit the superscript $t$ from $Q, K, V$ for brevity.}
    \label{fig:model}
\end{figure*}

We propose \proposed to address misalignment between semantically corresponding regions in the image pair. 
Specifically, we obtain sophisticated feature mapping between two images based on query-query alignment, and rearrange key and value for better query-key-value correspondence.
Lastly, refined attention is created by utilizing the rearranged keys and values, enabling the capture of semantic correspondence between both images.
The entire procedure of \proposed is illustrated in Figure~\ref{fig:model}.


\subsection{Query-Query alignment}\label{sec:qqalign}
The main cause of attention leakage is the feature misalignment in the cross-image attention. 
To mitigate this, we leverage the \textit{query-query alignment} between \( Q_{\text{str}} \) and \( Q_{\text{app}} \).
Since queries determine the semantic meaning of each spatial location in an image, we claim that query-query alignment can provide aligned features between two images more accurately than query-key alignment of Cross-Image. 
In this paper, we measure the query-query alignment with an alignment matrix \( S \):
\begin{equation}
 S = Q_{\text{app}} Q_{\text{str}}^\top,
\end{equation}
where $S \in \mathbb{R}^{n \times n}$ and $n$ is the number of queries.
Each element \( S_{r,c} \) indicates the similarity score between the \( r \)-th query of \( Q_{\text{app}} \) and the \( c \)-th query of \( Q_{\text{str}} \).

\vspace{0.1cm}
\noindent \textbf{Empirical Evidence.}
Figure~\ref{fig:attention_maps_comparison} shows that the query-query alignment (\(\text{softmax}(Q_{\text{str}} Q_{\text{app}}^\top)\)) achieves better correspondence than query-key alignment (\(\text{softmax}(Q_{\text{str}} K_{\text{app}}^\top)\)).
The query-query alignment provides more sophisticated spatial attention, while the query-key alignment distributes attention broadly across related regions.
For example, for the finial tip in \( I_{\text{app}} \) (second column in the green grid), query-key alignment distributes attentions to the body part of the building in \( I_{\text{str}} \), while our query-query alignment sharply focuses on the finial tip in \( I_{\text{str}} \).
Similarly, for the right pitched roof in \( I_{\text{app}} \) (third column in the green grid), query-key alignment broadly focuses on the background in \( I_{\text{str}} \), whereas our query-query alignment precisely attends to the right pitched roof area in \( I_{\text{str}} \).
Therefore, by leveraging the query-query alignment to extract precise and detailed points of interest, we could enable the model to accurately and comprehensively focus on relevant regions.

\subsection{Key-Value Rearrangement}\label{sec:keyre}
We utilize the query-query alignment $S$ between $Q_{\text{str}}$ and $Q_{\text{app}}$ to rearrange $(K_{\text{app}}, V_{\text{app}})$ for better query-key-value correspondence, based on our demonstration that $S$ provides precise feature alignment between \( I_{\text{str}} \) and \( I_{\text{app}} \).
It is important to recognize that we modify the keys and values rather than the queries, as queries define the semantic meaning of each spatial location in an image.
Modifying the queries could potentially disrupt the structural semantics of the output image.

\vspace{0.1cm}
\noindent \textbf{Rearrangement Matrix.}
Specifically, we construct an aggregation matrix \( P \in \mathbb{R}^{n \times n} \) for the key-value rearrangement based on the query-query alignment $S$, as follows:
\begin{equation}
(P^\top)_{r,c} = \begin{cases}
\frac{1}{k}, & \text{if } S_{r,c} \in \text{Top-}k (S_{r}) \\
0, & \text{otherwise}
\end{cases}
\end{equation}
$\text{Top-}k (S_{r})$ denotes the set of top-$k$ values in each row $S_{r} \in \mathbb{R}^n$.
It is worth noting that we only retain top-$k$ alignments for key-value rearrangement, as larger \( k \) broadens the range of features aggregated and may dilute the distinctiveness of the generated image.
In this paper, we use $k=1$.
If the sum of any row in \( P \) is 0, we set the diagonal entry to 1 to maintain the original keys/values:
\begin{equation}
\quad P_{r,r} = 1, \text{ If } \sum_{c=1}^n P_{r,c} = 0.
\end{equation}
We then apply row-wise softmax reweighting to \( P \):
\begin{equation}
P'_{r,c} = \begin{cases}
\frac{e^{P_{r,c}}}{\sum_{c' \in [n], P_{r,c'} \neq 0} e^{P_{r,c'}}}, & \text{if } P_{r,c} \neq 0 \\
0, & \text{if } P_{r,c} = 0
\end{cases}
\end{equation}
It is noted that we ensure that elements with an initial value of 0 remain unchanged.
Each element $P'_{r,c}$ indicates the contribution weights of the $c$-th feature on the $r$-th feature.

\vspace{0.1cm}
\noindent  \textbf{Key-Value Rearrangement via Aggregation.}
The aggregation matrix \( P' \) is applied to modify the original key matrix \( K_{\text{app}} \) and the original value matrix \( V_{\text{app}} \), to create a rearranged key matrix \( K^* \) and value matrix \( V^* \):
\begin{align}
    K^* &= P' K_{\text{app}},\\
    V^* &= P' V_{\text{app}}.
\end{align}
For example, if \( P'_{10,20} = 1 \), the 20th key and value are rearranged to be aligned with the 10th query.
By leveraging the query-query alignment to rearrange the key/value matrix, \( K^* \) and  \( V^* \) achieve improved feature alignment with \( Q_{\text{str}} \). This alignment extends to \( Q_{\text{out}} \) used for generating the output image.
Consequently, the modified key \( K^* \) and the modified value \( V^* \) align their features with the most relevant query vectors, thereby mitigating the issue of attention leakage.

\begin{figure*}[t]
\centering  
    \includegraphics[width=0.9\linewidth]{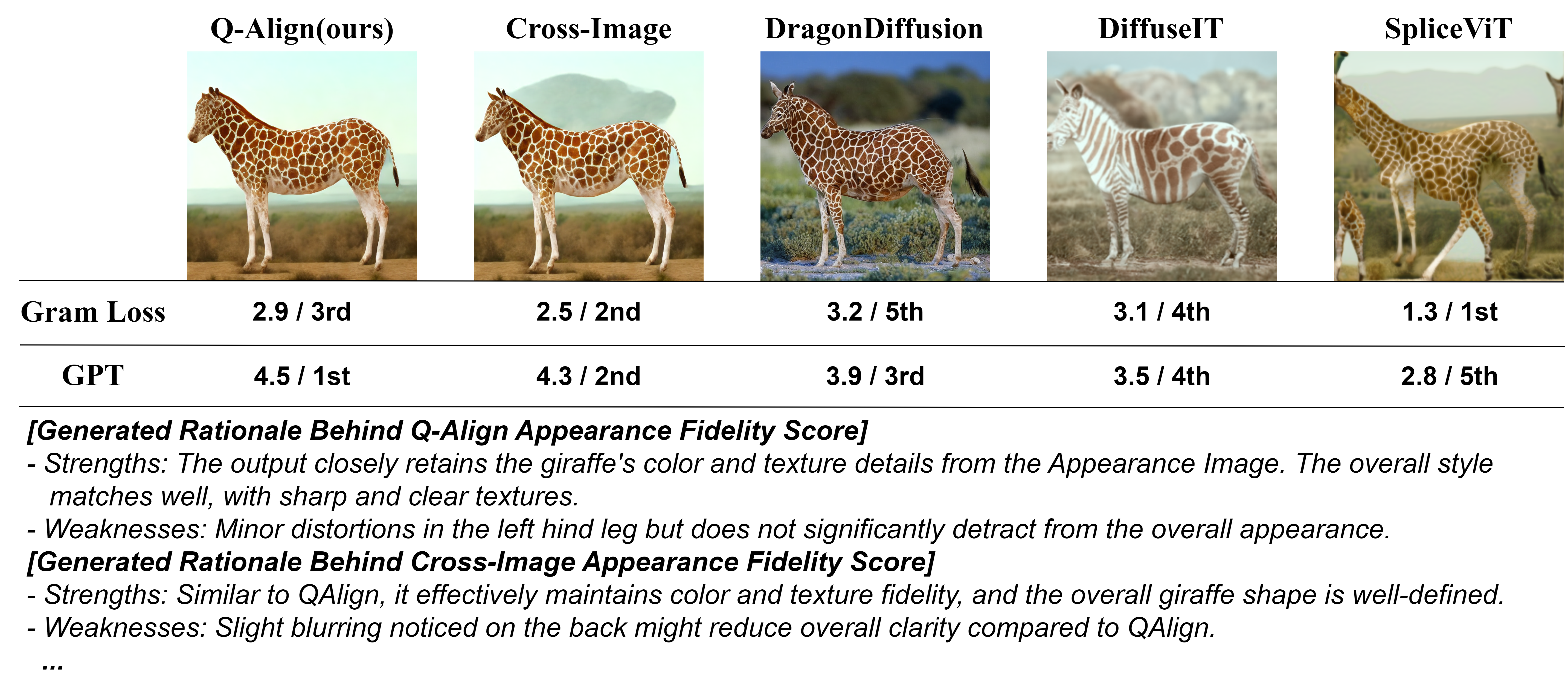}    
    
    \caption{Comparison between our GPT-based Appearance Fidelity Score (Table~\ref{table:appearance_combined}) and Gram Loss (Table~\ref{tab:category_mean_std_combined}). The example is from Figure~\ref{fig:qualitative}, and details can be found in the supplementary material.}
    \label{fig:gpt_eval}    
\end{figure*}
 
\vspace{0.1cm}
\noindent \textbf{Rearranged Cross-Image Attention}\label{sec:crossq}
Finally, we deploy the rearranged key \( K^* \) and value \( V^* \) for the Cross-Image attention between \( I_{\text{out}} \) and \( I_{\text{app}} \):
\begin{equation}
\text{softmax}\left(\frac{Q_{\text{out}} (K^*)^\top}{\sqrt{d}}\right) V^*.
\end{equation}
This formulation maintains semantic consistency during the \task, ensuring that \( Q_{\text{out}} \), \( K^* \), and \( V^* \) are aligned correctly while reducing the generation of unintended artifacts.

%% file: sections/4_experiment.tex
\section{Experiments}
\label{sec:experiment}

\subsection{Experiment setting}
\noindent \textbf{Datasets.}
Following Cross-Image~\cite{cross-image}, we selected \textit{six} domains (animal faces, animals, cars, birds, buildings, and cakes) to enable a fair comparison.
As done in Cross-Image \cite{cross-image}, for the animal face domain, images from AFHQ~\cite{AFHQ} were utilized.
For the remaining domains, we utilized images from the official Cross-Image page as well as copyright-free real-world images from Unsplash\footnote{\url{https://unsplash.com}}. To ensure diversity in the dataset, we gathered 30 image pairs per domain, resulting in a total of 180 pairs.

\vspace{0.1cm}
\noindent  \textbf{Baseline Methods.}
We compare {\proposed} with state-of-the-art approaches, including Cross-Image~\cite{cross-image}, DragonDiffusion~\cite{dragondiffusion}, DiffuseIT~\cite{diffuseit}, and SpliceViT~\cite{splice}.
All methods were implemented using the official code provided by the respective authors, following their default parameters.
We use Stable Diffusion~\cite{stablediffusion} as the base model, and the same random seed and other processing parameters are applied uniformly.
Further details on parameter settings and implementation specifics are provided in the supplementary materials.

\begin{table}[t]
\centering
\scriptsize
\renewcommand{\arraystretch}{1.1}
\resizebox{0.5\textwidth}{!}{  
\begin{tabular}{|c|cc|cc|}
\hline
\multirow{2}{*}{Domain} & \multicolumn{2}{c|}{Gram Loss$\downarrow$} & \multicolumn{2}{c|}{IoU$\uparrow$} \\
\cline{2-5}
                        & Cross-Image & Q-Align         & Cross-Image & Q-Align \\
\hline
Animal       & 1.08 / 0.64 & 1.68 / 1.77 & 0.66 / 0.16 & 0.71 / 0.12 \\
Animal Face  & 0.79 / 0.43 & 0.84 / 0.42 & 0.72 / 0.18 & 0.78 / 0.10 \\
Bird         & 1.47 / 0.85 & 1.37 / 0.75 & 0.75 / 0.15 & 0.80 / 0.09 \\
Building     & 4.92 / 4.03 & 4.68 / 2.38 & 0.81 / 0.17 & 0.82 / 0.17 \\
Cake         & 3.16 / 3.26 & 2.58 / 2.11 & 0.78 / 0.23 & 0.84 / 0.11 \\
Car          & 1.12 / 0.32 & 1.32 / 0.35 & 0.84 / 0.17 & 0.83 / 0.17 \\
\hline
Avg          & 2.09 / 1.59 & \textbf{2.08} / 1.30 & 0.76 / 0.18 & \textbf{0.80} / 0.13 \\
\hline
\end{tabular}
}
\caption{Mean and standard deviation of Gram Loss and IoU for each domain (Mean / Std).}
\label{tab:category_mean_std_combined}
\end{table}

\begin{table}[t]
\centering
\begin{minipage}[t]{0.495\textwidth}
    \centering
    \resizebox{\textwidth}{!}{
        \begin{tabular}{|c|c|c|c|c|c|}
            \hline
            \multicolumn{6}{|c|}{Appearance Fidelity Score} \\
            \hline
            Domain & SpliceViT & DiffuseIT  & DragonDiffusion & Cross-Image & Q-Align (ours) \\
            \hline
            Animal       & 2.54 / 0.51 & 3.65 / 0.47 & 3.69 / 0.58 & 4.04 / 0.50 & 4.25 / 0.41 \\
            Animal Face  & 2.49 / 0.58 & 3.41 / 0.39 & 3.72 / 0.53 & 4.29 / 0.28 & 4.31 / 0.30 \\
            Bird         & 2.68 / 0.39 & 3.36 / 0.48 & 3.88 / 0.53 & 4.15 / 0.29 & 4.31 / 0.19 \\
            Building     & 2.16 / 0.50 & 3.20 / 0.39 & 3.52 / 0.35 & 4.36 / 0.14 & 4.46 / 0.17 \\
            Cake         & 2.56 / 0.64 & 2.93 / 0.49 & 4.15 / 0.51 & 4.09 / 0.42 & 4.23 / 0.32 \\
            Car          & 2.23 / 0.40 & 3.21 / 0.40 & 3.86 / 0.41 & 4.22 / 0.36 & 4.21 / 0.42 \\
            \hline
            Avg          & 2.44 / 0.54 & 3.29 / 0.49 & 3.80 / 0.53 & \underline{4.19} / 0.37 & \textbf{4.30} / 0.33 \\
            \hline
        \end{tabular}
    }
    \caption{GPT-based appearance fidelity score across different methods (Mean/Std).}
    \label{table:appearance_combined}
\end{minipage}
\hfill
\begin{minipage}[t]{0.495\textwidth}
    \centering
    \resizebox{\textwidth}{!}{
        \begin{tabular}{|c|c|c|c|c|c|}
            \hline
            \multicolumn{6}{|c|}{Structural Consistency Score} \\
            \hline
            Domain & SpliceViT & DiffuseIT  & DragonDiffusion & Cross-Image & Q-Align (ours) \\
            \hline
            Animal       & 2.66 / 0.50 & 3.77 / 0.57 & 3.76 / 0.71 & 3.93 / 0.57 & 4.17 / 0.46 \\
            Animal Face  & 2.54 / 0.50 & 3.42 / 0.30 & 3.73 / 0.55 & 4.35 / 0.29 & 4.41 / 0.22 \\
            Bird         & 2.81 / 0.51 & 3.41 / 0.48 & 4.00 / 0.58 & 4.08 / 0.42 & 4.21 / 0.34 \\
            Building     & 2.31 / 0.55 & 3.23 / 0.41 & 3.42 / 0.48 & 4.41 / 0.19 & 4.50 / 0.14 \\
            Cake         & 2.85 / 0.67 & 3.02 / 0.63 & 4.26 / 0.57 & 4.15 / 0.36 & 4.24 / 0.29 \\
            Car          & 2.34 / 0.34 & 3.28 / 0.38 & 4.07 / 0.44 & 4.33 / 0.22 & 4.32 / 0.29 \\
            \hline
            Avg          & 2.58 / 0.56 & 3.35 / 0.53 & 3.87 / 0.62 & \underline{4.21} / 0.40 & \textbf{4.31} / 0.33 \\
            \hline
        \end{tabular}
    }
    \caption{GPT-based structural consistency score across different methods (Mean/Std).}
    \label{table:structural_combined}
\end{minipage}
\end{table}

\begin{figure*}[t]
    \centering  
    \includegraphics[width=0.9\linewidth]{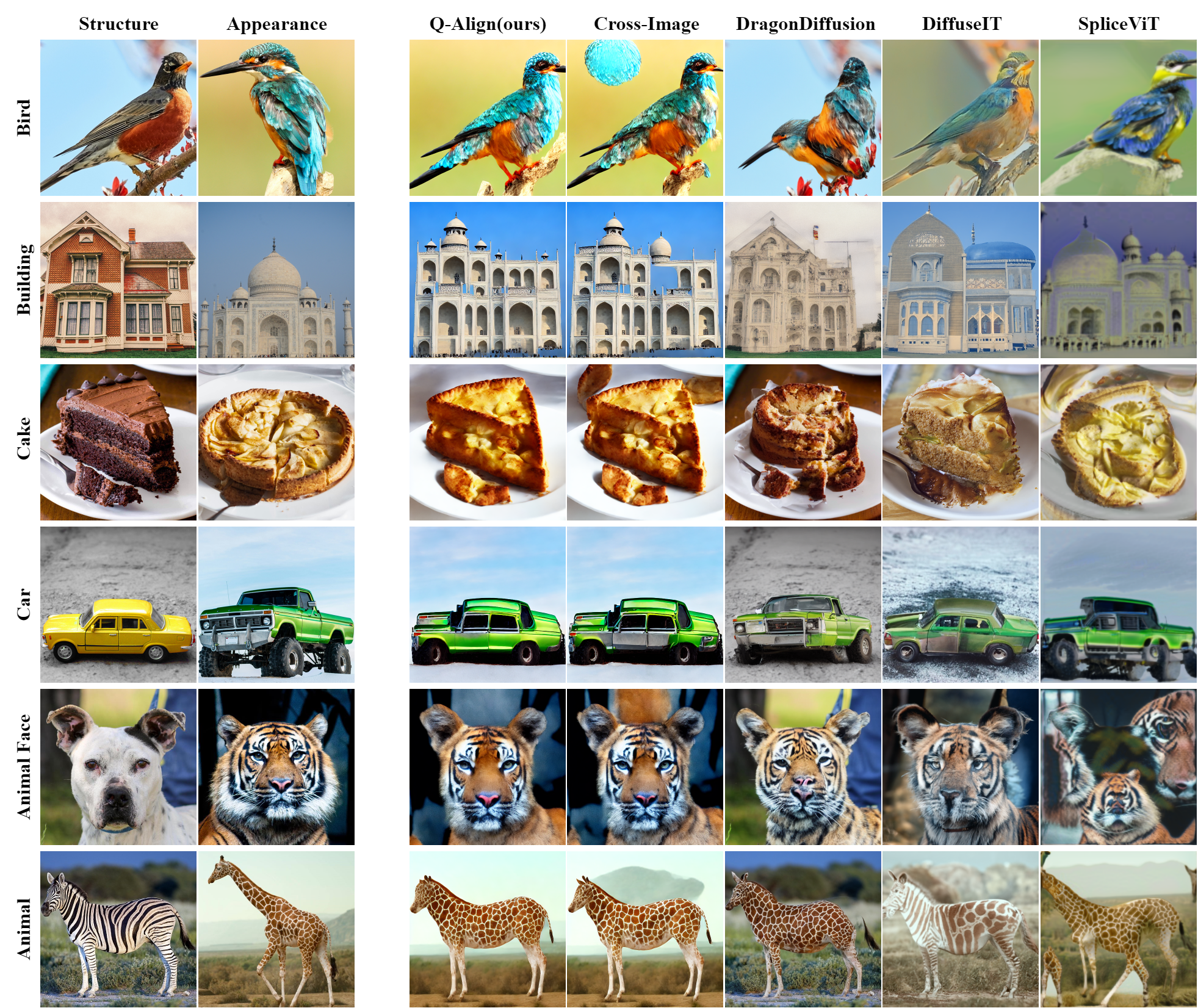}    
    
    \caption{Qualitative comparison across different methods. Each row displays an example consisting of the structure image, appearance image, and the outputs from our method, Cross-Image, DragonDiffusion, DiffuseIT, and SpliceViT. For demonstration purposes, we select samples that exhibit attention leakage in Cross-Image. More qualitative results are presented in the supplementary materials.}
    \label{fig:qualitative}   
\end{figure*}

\vspace{0.1cm}
\noindent  \textbf{Evaluation Metrics (existing).}
In the earlier work \cite{cross-image}, they used Gram loss~\cite{grammatrix} computed on VGG-19 features~\cite{vgg} for the appearance fidelity, and IoU between object masks~\cite{HQ-SAM} of \( I_{\text{out}} \) and \( I_{\text{str}} \) for the structural consistency.
However, as shown in Table~\ref{tab:category_mean_std_combined}, applying those evaluation metrics proposed in Cross-Image led to significant variance across different samples, which limited the reliability of the quantitative assessment. 
Furthermore, Figure~\ref{fig:gpt_eval} shows that the appearance similarity perceived by humans does not align well with the results of the Gram loss metric.

\vspace{0.1cm}
\noindent  \textbf{GPT-based Metrics (proposed).}
To address these limitations, we propose a novel evaluation protocol leveraging GPT-4o series \cite{gpt-4o}\footnote{we use gpt-4o-mini-2024-07-18 with OpenAI API}, a powerful large language model with a vision capability.
Inspired by \cite{geval}, we prompt GPT with a detailed explanation of the appearance transfer task and explicitly defined two evaluation criteria (\textit{i.e.}, appearance fidelity and structural consistency).
We then instruct GPT to rate each criterion on a scale from 1.0 to 5.0. 
A comprehensive description of the prompt design is provided in the supplementary material.
As shown in Figure~\ref{fig:gpt_eval}, GPT assigns scores at a level comparable to human evaluation while also providing explanations for its reasoning.

\subsection{Experiment results}

\noindent \textbf{Quantitative Comparison.}
Table~\ref{table:appearance_combined} and Table~\ref{table:structural_combined} present the quantitative evaluation of \task across different methods, based on the proposed GPT-based appearance fidelity and structural consistency scores.
In terms of appearance, \proposed achieved the best results in most domains, with the highest average score.
In terms of structure, \proposed also achieved the highest average structural consistency score, surpassing other methods by notable margins.
By addressing the attention leakage issue and preventing unnecessary artifacts, {\proposed} achieved both strong structure preservation and high appearance similarity, indicating that it delivered the best overall results for the \task task.

\begin{figure}[t]
    \centering  
    \includegraphics[width=0.9\linewidth]{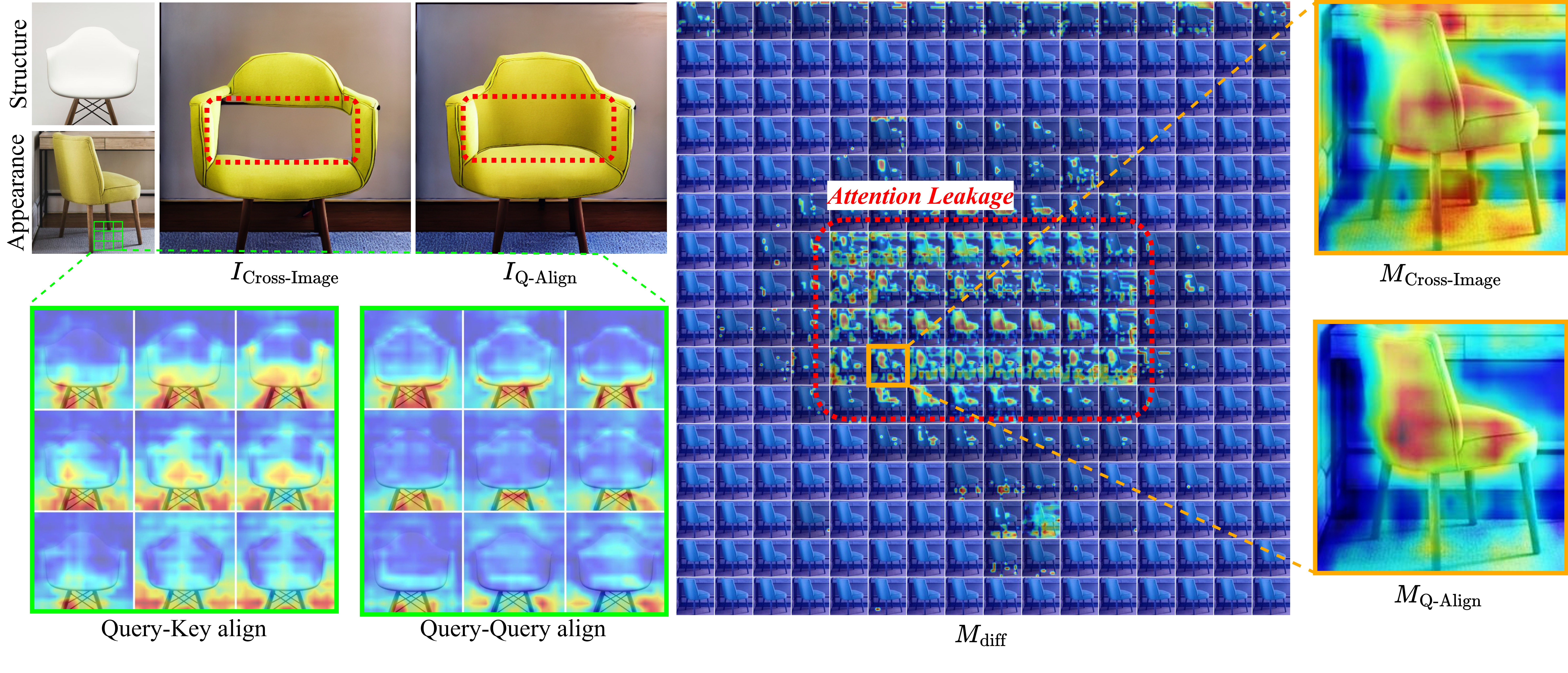}    
    
    \caption{Zero-shot \task with Cross-Image and \proposed. We present Query-Key alignment and Query-Query alignment for the 3*3 green grid. \( M_{\text{diff}} = | M_{\text{Cross-Image}} - M_{\text{\proposed}} | \) represents attention difference. Values below 0.2 are masked as 0 to emphasize only significant differences.}
    \vspace{-0.2cm}
    \label{fig:leakge_analysis_figure}   
\end{figure}


\vspace{0.1cm}
\noindent \textbf{Qualitative Results.}
Figure~\ref{fig:qualitative} presents a comparison of \task results across different models, highlighting the distinct advantages of {\proposed}.
DragonDiffusion showed vibrant color but produced distorted images. 
As a mask-based approach, it prioritized fitting the style within the mask region rather than ensuring accurate semantic mapping. 
For instance, in the building example, the appearance image contained rounded roofs, meaning that the appearance transfer output should exhibit a rounded roof shape.
However, DragonDiffusion forced the style into the original pointed roof mask, resulting in an unnatural composition.
Cross-Image exhibited attention leakage, where objects appeared in the background, background elements were incorrectly mapped onto objects, or incorrect semantic mappings occurred within both object and background regions. 
For example, in the bird image, wing characteristics appeared in the background, and in the vehicle image, bumper traits (e.g., gray tones) were mapped onto the doors. 
Additionally, in the animal category, the background color was incorrectly mapped.
In contrast, {\proposed} demonstrated reliable performance in preserving both structural and appearance features, consistently generating high-quality and faithful \task results across diverse examples.

\subsection{Study on \proposed}
Figure~\ref{fig:leakge_analysis_figure} shows the \task results of Cross-Image and \proposed.
We first observe that Cross-Image generates an undesirable image of a chair.
A clue for this issue can be found in the Query-Key alignment, where the background is incorrectly aligned to the center of the chair. 
As a result, Cross-Image generates a background-colored hole in the chair.
Query-Query alignment, on the other hand, prevents any attention leakage from the background onto the chair.
The attention difference map \( M_{\text{diff}} \) highlights areas with significant differences in semantic focus, emphasizing central regions where attention leakage occurs in Cross-Image.
We observe that, when generating the center of the chair, Cross-Image attends to both the chair and the background, whereas Q-Align focuses solely on the chair itself.
This result aligns with Zhang et al.~\cite{attention_alignment}, who highlighted that queries and keys reside in distinct embedding spaces.
We assert that query-query alignment, occurring within the same embedding space, inherently achieves better alignment than query-key alignment.

%% file: sections/5_related.tex
\section{Related Work}

\noindent  \textbf{Semantic Alignment.}
Most studies have focused on learning semantic correspondences between images through supervised training~\cite{patchmatch, cat, gocor, raft}. 
While these approaches demonstrate high performance, they come with the limitation of requiring large-scale datasets. 
To address this, methods that utilize features from pre-trained vision transformers~\cite{dino, vit, asic} have been proposed, along with the image manipulation methodologies~\cite{splice, diffuseit}.
More recent approaches utilize large-scale image generation models~\cite{stablediffusion,vqdiffusion,dalle} to extract features that enable precise semantic mapping between images~\cite{DIFT, diffusion_semantic_hyperfeature, diffusion_semantic_tale_two_feature, diffusion_semanticunsupervised}. 
Additionally, controlling the attention map of large models for manipulation has been explored~\cite{masactrl,p2p,pix2pix_zero,plug_and_play}, enabling semantic coherence to be preserved without the need for additional training.

\vspace{0.1cm}
\noindent \textbf{Attention in Diffusion Models.}
Attention mechanisms~\cite{attention} are fundamental in diffusion-based image manipulation, with numerous studies focused on optimizing~\cite{DPL}, modifying~\cite{masactrl, inversion_free, flexiedit,p2p,pix2pix_zero}, and refining~\cite{ZONE, diffedit}. 
Among these, methods that mix the query of one image with the key and value of another have been developed to support non-rigid image editing~\cite{masactrl, inversion_free, flexiedit}.
Other studies utilize refined attention masks to guide the accurate designation of regions for modification within an image~\cite{ZONE, diffedit}, while some employ latent optimization to prevent focusing on irrelevant regions~\cite{DPL}.
In this study, we propose a zero-shot method that operates without any external masks or additional optimization, relying solely on query-query alignment. 
This design effectively mitigates attention leakage while preserving both structural consistency and semantic fidelity unlike mask-based approaches that distort semantics by forcing styles into fixed mask regions.


%% file: sections/6_conclusion.tex
\section{Conclusion}
We proposed Q-Align, a novel method that utilizes Query-Query alignment to effectively rearrange keys and values for the attention mechanism.
Our method offers a zero-shot, straightforward solution that directly addresses the critical issue of attention leakage observed in SOTA methods.
Through comprehensive analysis and extensive experiments, we demonstrated that Q-Align not only effectively resolves attention leakage but also significantly outperforms existing approaches.
As future work, we envision extending Q-Align to incorporate fine-grained semantic mapping techniques, which we expect will further enhance its capability for precise and high-quality appearance transfer.

%% file: sections/7_acknowledgement.tex
\section*{Acknowledgments}
This work was supported by the Digital Innovation Hub project supervised by the Daegu Digital Innovation Promotion Agency (DIP), grant funded by the Korea government (MSIT and Daegu Metropolitan City) in 2025 (No.~25DIH-11, Development of Model Context Protocol (MCP)-Based Multi-Agent Collaborative System for Large Language Models (LLMs)). 
※~MSIT: Ministry of Science and ICT.